\documentclass[letterpaper, 10 pt, conference]{ieeeconf}  % Comment this line out if you need a4paper

\usepackage{graphicx} 
\usepackage{booktabs}
\usepackage{array}
\usepackage{multirow}
\usepackage{amsmath}

\IEEEoverridecommandlockouts                              % This command is only needed if 
\title{\LARGE \bf
Heuristic Search for Structural Constraints in Data Association
}

\author{Xiao Zhou$^{1}$, Peilin Jiang$^{2}$ and Fei Wang$^{1}$% <-this % stops a space
\thanks{$^{1}$Xiao Zhou and Fei Wang are with the Institute of Artificial Intelligence \& Robotics, Xi’an Jiaotong University, Xi'an, China
        {\tt\small zx2962@stu.xjtu.edu.cn,wfx@mail.xjtu.edu.cn}}%
\thanks{$^{2}$Peilin Jiang is with the School of Software Engineering, Xi’an Jiaotong University, Xi'an, China
        {\tt\small pljiang@mail.xjtu.edu.cn}}%
}

\begin{document}

\maketitle
\thispagestyle{empty}
\pagestyle{empty}

%%%%%%%%%%%%%%%%%%%%%%%%%%%%%%%%%%%%%%%%%%%%%%%%%%%%%%%%%%%%%%%%%%%%%%%%%%%%%%%%
\begin{abstract}

The research on multi-object tracking (MOT) is essentially to solve for the data association assignment, the core of which is to design the association cost as discriminative as possible. Generally speaking, the match ambiguities caused by similar appearances of objects and the moving cameras make the data association perplexing and challenging. In this paper, we propose a new heuristic method to search for structural constraints (HSSC) of multiple targets when solving the problem of online multi-object tracking. We believe that the internal structure among multiple targets in the adjacent frames could remain constant and stable even though the video sequences are captured by a moving camera. As a result, the structural constraints are able to cut down the match ambiguities caused by the moving cameras as well as similar appearances of the tracked objects. The proposed heuristic method aims to obtain a maximum match set under the minimum structural cost for each available match pair, which can be integrated with the raw association costs and make them more elaborate and discriminative compared with other approaches. In addition, this paper presents a new method to recover missing targets by minimizing the cost function generated from both motion and structure cues. Our online multi-object tracking (MOT) algorithm based on HSSC has achieved the multi-object tracking accuracy (MOTA) of 25.0 on the public dataset 2DMOT2015 \cite{leal2015motchallenge}.

\end{abstract}

%%%%%%%%%%%%%%%%%%%%%%%%%%%%%%%%%%%%%%%%%%%%%%%%%%%%%%%%%%%%%%%%%%%%%%%%%%%%%%%%
\section{INTRODUCTION}

Currently, researches on MOT algorithms mainly focus on the problem of data association with given detections, and tracking-by-detection methods can be generally classified into online and offline MOT methods. For the offline MOT methods, attentions are often paid to the association assignment between tracklets, which would inevitably utilize detections from future frames. Therefore, offline MOT methods are considered as a global optimization scheme, and it can only be applied to non-real-time occasions such as offline video analysis. On the other hand, online MOT methods focus on the data association between detections in the current frame and historical tracklets, which do not need detections from future frames as inputs and thus can be applied to real-time situations such as video surveillance, robot navigation and so forth.

\begin{figure}[thpb]
      \centering
      \graphicspath{ {images/} }     
      \includegraphics[scale=0.45]{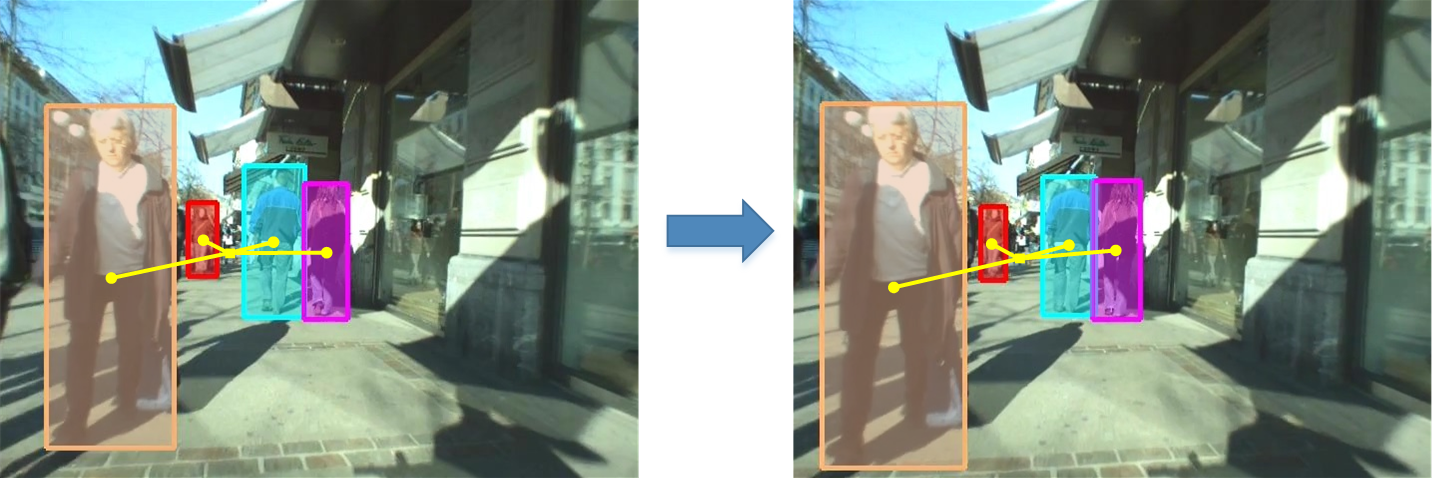}
      \caption{Structural constraints of multiple targets between adjacent frames. The pictures are consecutively extracted from the video frames in the dataset ETH-Sunnyday \cite{leal2015motchallenge}, which is captured by a moving camera. The yellow solid squares denote the center location of all pedestrian targets, and the yellow lines with rounded ends represent the internal structures among the four pedestrian targets. Obviously, the structures in both images basically remains Invariable, which inspires us to search for structural constraints among multiple targets in data association assignment.}
      \label{figurelabel1}
   \end{figure}

The design of pairwise association costs between detections and historical tracklets directly affects the performance of the data association assignment. In MOT methods, appearance features and motion cues of targets are usually extracted to obtain pairwise association costs. As the appearance models can help distinguish between different objects very well in most circumstances, many studies are committed to finding more discriminative appearance features, such as \cite{solera2015learning}, \cite{bewley2016alextrac}, \cite{bae2014robust}. However, the pairwise cost acquired from appearance information would be easily unreliable when tracking similar appearance targets. Therefore, the movement information becomes necessary to discriminate and identify multiple targets. It is undeniable that, in a fixed scene, we can use a linear motion model to characterize a simple moving target’s trajectory or establish an autoregressive motion model \cite{dicle2013way} for a complex one. However, for a moving camera, the movement of the individual target becomes the superposition of both the object’s movement itself and the translation of the frame image due to the moving camera. Under such circumstances, observable motion cues become unstable, unpredictable and unreliable. The conventional motion models can hardly describe the movement features of these targets, let alone making accurate prediction in the next frame. Although the uncertainty of the target motion deteriorates with the moving camera, the internal structure among multiple targets in the adjacent frames can approximately remain consistent and steady, which could be utilized to compensate the short board on account of ambiguous motion cues, as is shown in figure \ref{figurelabel1}. 

In this paper, we propose a heuristic approach to search for internal structural constraints between multiple targets to modify pairwise cost matrix and thus lessen the association ambiguities in the data association assignment. Furthermore, we propose a new method based on minimization of the cost function constructed by both motion and structure cues. This is applied to predict the next location of a missing target and thus it can be associated with the reappearing target. This proposed method based online MOT algorithm primarily consists of two steps. First, we construct the association cost matrix between the detections in the current frame and the historical target tracklets, including the design of the raw pairwise costs and the amelioration by the structural constraints. In specific, we exploit motion and appearance cues to construct the raw pairwise costs and then use the proposed heuristic approach to search for the optimal structural constraints to ameliorate the association cost matrix. The second step is the association assignment, in which we utilize generalized linear assignment \cite{dicle2013way} to match the available detections with the historical tracklets, and then recover the missing targets in a fixed size window by minimizing the cost function constructed by both motion and structure cues.

\section{RELATED WORK}

%\subsection{Selecting a Template (Heading 2)}

We review related MOT methods that pay much attention to motion cues and the structure of multiple targets. Andriyenko et al. \cite{andriyenko2011multi}, \cite{milan2014continuous}, \cite{milan2016multi} focus on designing an energy function and constructing an optimization scheme to find local minima of the energy function. For this method, all the detection responses are the input of the energy function, the solution space of which contains all possible associations. The velocity models are used in this method to descript the motion of targets, which can only cover the situation of simple movements.
 
Dicle et al. \cite{dicle2013way} only use motion cues to track multiple objects with similar appearance, the autoregressive model is utilized to represent the motion of each target and to construct association cost. This model handles complex target movements and similar appearances, but cannot work well when faced with non-stationary cameras. Possegger et al. \cite{possegger2014occlusion} exploit the geometric information, including occlusion information, detector reliability, and motion prediction, to recover missing objects. Collins et al. \cite{collins2012multitarget} develop a higher-order cost function for data association. This method uses active contour spline energy to measure the quality of a proposed trajectory. Liu et al. \cite{liu2013tracking} extract game context features from noisy detections to build context-conditioned motion models for tracking sports players. Yang and Nevatia \cite{yang2012online} use online conditional random field (CRF) to produce unary and pairwise energy functions based on linear and smooth motion to solve the multi-object tracking problems. Other trackers, like \cite{milan2015joint}, \cite{kim2016cdt}, \cite{tang2014detection} joint both the detection and the tracking assignments.

Yoon et al. \cite{yoon2015bayesian} utilize the motion context to construct a relative motion network to deal with the unexpected camera motions. However, this method cannot handle abrupt camera motions and fluctuations. In \cite{hong2016online}, Yoon et al. exploit the structural motion constraints and propose an event aggregation approach to solve the MOT problem with moving cameras. This method shows great performance in public datasets. However, in this approach, motion cues of the targets only depend on their relative structure and the update of structure relies on a linear motion model, which can hardly handle the objects with sophisticated movement or high-speed motion.

The MOT problem can be regarded as an association assignment between the object detections and historical tracklets for online trackers, or tracklets-to-tracklets for offline trackers. There are about two directions to develop a tracker and make it more sophisticated. On the one hand, many researches focus on the tool of solving the data association. Therefore, many different approaches are adopted to solve the association assignment, including minimization of network flow based cost  \cite{dehghan2015target} \cite{chari2015pairwise}, \cite{butt2013multi}, \cite{haubold2016generalized}, linear programming \cite{dicle2013way}, \cite{jiang2007linear}, Hungarian matching \cite{hong2016online}, \cite{bewley2016alextrac}, \cite{huang2008robust} and subgraph decomposition \cite{tang2015subgraph},\cite{dehghan2015gmmcp}. On the other hand, design of features that are more elaborate \cite{zhang2016tracking}, \cite{bae2014robust} and making association cost more discriminative have been broadly concerned. For example, Rezatofighi et.al. \cite{hamid2015joint} modify the association costs with joint probabilities, and decompose the original problem into a series of integer programming, which is more efficient and time-saving. However, in the method, each pairwise cost is the aggregation of costs generated by all possible match event under the fixed pair, and thus the most likely associated pair would be susceptible to noise caused by almost impossible associated pair, especially when detections and targets cannot match each other one-to-one. So does the tracker in \cite{hong2016online}.

%\subsection{Maintaining the Integrity of the Specifications}

\section{PROPOSED METHOD}

\subsection{Overview} The discrimination of pairwise costs is of great significance for solving the association assignments. In the section of related work, a variety of cost design methods are discussed, most of which are concerning about objects’ appearance and motion cues. For example, \cite{hong2016online} proposes an event aggregation method to integrate structural constraints in all possible assignment events. In this paper, we propose a heuristic method to search for the optimal association event corresponding to the minimum structural cost for each possible pair.

\subsection{Heuristic search for the optimal structural constraint}

\begin{figure*}[thpb]
      \centering
      \graphicspath{ {images/} }     
      \includegraphics[scale=0.93]{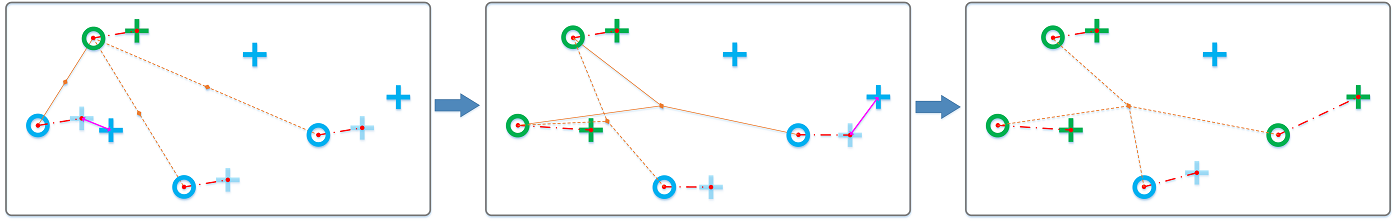}
      \caption{Heuristic search for structural constraints. The circles represent the targets in the last frame, the opaque crosses denote detections in the current frame, and the translucent crosses represent the optimal prediction of the targets under the current structural constraint by using Eq. (\ref{eq4}). There are two states of detections or targets, including the in-match-set with green marker, and the out-of-match-set with blue marker. The orange dash lines express the alternative structural constraints between the match set and other available pairs. The orange solid lines indicate the optimal structural constraints under the current condition. In the first graph, the match set is initialized as the given associated pair marked in green. Using the Eq. (\ref{eq4}), we can obtain the optimal prediction of each target in the current frame, which is denoted by translucent crosses. Then it would be easy to find a detection coupled with the nearest target that minimize the structural cost between detections and targets in both the match set and the available pair. The second graph shows the cost is less than the threshold, and thus the available pair is merged into the match set. In the third graph, we can see the algorithm tries to find the forth element of the match set, but the structural cost is far beyond the cost threshold. Therefore, the heuristic search is terminated and the output is the match set with three pairs.}
      \label{figurelabel2}
   \end{figure*}

The consecutive video sequences guarantee the continuity of the objects’ motion, thus making it possible to exploit structure cues between multiple targets in a frame. We denote the center location of a detection $i$ at frame $t$ as $d_{i}^{t}(x,y)$ and an object $j$ at frame $t-1$ as $T_{j}^{t-1}(x,y)$ (In the subsequent formulas, we will omit $(x,y)$ ). We use $\Delta d_{i}^{t}$ to denote the relative displacement between the $i$th object location and the spatial distribution center of all detection positions, which is as shown as follows.

\begin{equation}
\Delta d_{i}^{t}=d_{i}^{t}-\sum\limits_{k=1}^{n}{d_{k}^{t}},\Delta T_{j}^{t-1}=T_{j}^{t-1}-\sum\limits_{k=1}^{m}{T_{k}^{t-1}}
\label{eq1}
\end{equation}

If the detection $i$ in the frame $t$ is associated with the object $j$, the assignment is denoted by $\{ {{X}_{ij}}\text{=1}\}$ or $\{d_{i}^{t},T_{j}^{t-1}\}$. Otherwise, it is denoted by $\{ {{X}_{ij}}\text{=0}\}$. Assuming there are $m$ detections in the current frame and $m$ historical objects in the past, and they could also match each other one-to-one, then the optimal assignment solution based on minimum structural cost of each possible pair $\{ {{X}_{pq}}\text{=1}\}$ can be constructed as:

\begin{equation}
\begin{aligned}
  & \widehat{{{X}_{ij}}}\text{=}\arg \underset{{{X}_{ij}}}{\mathop{\min }}\,\sum\limits_{i=1,i\ne p}^{m}{\sum\limits_{j=1,j\ne q}^{m}{{}}} \\ 
 & {{\left\| \Delta d_{i}^{t}-\Delta T_{j}^{t-1} \right\|}^{2}}{{X}_{ij}}+{{\left\| \Delta d_{p}^{t}-\Delta T_{q}^{t-1} \right\|}^{2}} \\ 
\end{aligned}
\label{eq2}
\end{equation}

Obviously, the above formula can be easily solved by the Hungarian algorithm. However, the actual number of detections in the current frame are usually different from the number of objects in the past due to false positives, false negatives and other noises. In other words, it can neither find global structural constraints between all detections and objects nor promise one-to-one match. Thus, Eq. (\ref{eq2}) cannot be used to fit the practical assignments. As such, we present a heuristic method to search for the optimum match set by minimizing structural cost in each search loop, which is described as below.

With a certain association pair $\left\{ d_{m}^{t},T_{n}^{t-1} \right\}$ and the corresponding match set ${{\Theta }_{mn}}$, which is initialized as $\left\{ d_{m}^{t},T_{n}^{t-1} \right\}$, we are able to search alternative match pairs for the optimal pair $\left\{ d_{i}^{t},T_{j}^{t-1} \right\}$ by minimizing the structural cost function generated among elements in the match set and the alternative pair heuristically and continually. The structural cost function is described by:

\begin{equation}
\begin{aligned}
  & \left\{ \widehat{d_{i}^{t}},\widehat{T_{j}^{t-1}} \right\}\text{=}\arg \underset{\left\{ d_{i}^{t},T_{j}^{t-1} \right\}\notin {{\Theta }_{mn}}}{\mathop{\min }}\, \\ 
 & {{\left\| \Delta d_{i}^{t}-\Delta T_{j}^{t-1} \right\|}^{2}}+\sum\limits_{\left\{ p,q \right\}\in {{\Theta }_{mn}}}{{{\left\| \Delta d_{p}^{t}-\Delta T_{q}^{t-1} \right\|}^{2}}} \\ 
\end{aligned}
\label{eq3}
\end{equation}

If the structural cost computed from the above formula is smaller than a certain threshold ${{\varphi }_{\text{s}}}$, we add the pair $\left\{ d_{i}^{t},T_{j}^{t-1} \right\}$ into the matching set ${{\Theta }_{mn}}$ and repeat the entire procedure to find the next optimal pair on the circumstance of the updated match set. Otherwise, we end the search. The maximum match set ${{\Theta }_{mn}}$ has been found under such structural constraints.

When solving Eq.(\ref{eq3}), we fix the target and assume that the solution space of detections $\widehat{d_{i}^{t}}$ is consecutive. The extreme point under such condition appears as the following formula:

\begin{equation}
\widehat{d_{i}^{t}}=T_{j}^{t-1}+\frac{1}{n}\sum\limits_{\left\{ p,q \right\}\in {{\Theta }_{mn}}}{(d_{p}^{t}-T_{q}^{t-1})}
\label{eq4}
\end{equation}

Evidently, there exists a unique location prediction in the current frame for each historical object under the condition of the known match set. Due to the discretization of the detections, we have to find the closest detection with the solution $\widehat{d_{i}^{t}}$ of each target and extract the one that generates the minimum structural cost of the map between detections and targets in the current match set. If the minimum structural cost corresponding to the optimal pair is less than the certain threshold, we expand the match set. Otherwise jump out of the loop and end the search. The match set ${{\Theta }_{mn}}$ is what we need. The specific procedure is shown in Algorithm 1 and Fig. \ref{figurelabel2}.

\begin{table}[h]
\begin{tabular}{lcl}
\toprule
\textbf{Algorithm 1} Heuristic search for structural constraints \\
\midrule
\textbf{Input:} $\left\{ d_{m}^{t},T_{n}^{t-1} \right\}$
\hfill  $\triangleright$ a certain fixed association pair\\
\textbf{Input:} $d_{i}^{t}$ \hfill $\triangleright$ the detection $i$ in frame $t$\\
\textbf{Input:} $T_{j}^{t-1}$      \hfill      $\triangleright$ the target $j$ in frame $t-1$\\
\textbf{Output:} ${{\Theta }_{mn}}$  \hfill   $\triangleright$ the optimal match set for each fixed pair\\ 
\textbf{function}($\left\{ d_{m}^{t},T_{n}^{t-1} \right\}$,$d_{i}^{t}$,$T_{j}^{t-1}$)\\
\qquad ${{\Theta }_{mn}}\text{=}\left\{ d_{m}^{t},T_{n}^{t-1} \right\}$;$C_{structure}^{\min }\text{= 0}$;\\
\qquad \textbf{While} $count<n_{t}$(the number of targets)\\
\qquad \qquad Candidate pair set $P=\emptyset$\\ 
\qquad \qquad $count=count+1$\\
\qquad \qquad \textbf{for} $j=1:{n_{t}}$\\
\qquad \qquad \quad \ Find the closest detection $d_{i}^{t}$ with the location $\widehat{d_{i}^{t}}$ \\
\qquad \qquad \quad \ computed by Eq. (\ref{eq4})\\
\qquad \qquad \quad \ $P=P\cup \left\{ d_{i}^{t},T_{j}^{t-1} \right\}$\\
\qquad \qquad \textbf{end}\\
\qquad \qquad Compute minimization of structural cost from the alternative \\
\qquad \qquad pairs by $C_{structure}^{\min }=\min\limits_{\left\{ d_{i}^{t},T_{j}^{t-1} \right\}\in P},$\\

\qquad \qquad \qquad \ ${{{\left\| \Delta d_{i}^{t}-\Delta T_{j}^{t-1} \right\|}^{2}}+\sum\limits_{\left\{ p,q \right\}\in {{\Theta }_{mn}}}{{{\left\| \Delta d_{p}^{t}-\Delta T_{q}^{t-1} \right\|}^{2}}}}$\\
\qquad \qquad \textbf{if} $C_{structure}^{\min }<{{\varphi }_{\text{s}}}$\\
\qquad \qquad \quad ${{\Theta }_{mn}}\text{= }{{\Theta }_{mn}}\cup {{\left\{ d_{i}^{t},T_{j}^{t-1} \right\}}_{C_{structure}^{\min }}}$;\\
\qquad \qquad \textbf{else}\\
\qquad \qquad \quad \textbf{break;}\\
\qquad \qquad \textbf{end}\\
\qquad \textbf{end}\\
\textbf{end}\\
\bottomrule
\end{tabular}
\label{algorithm}
\end{table}

\subsection{Pairwise cost}

In this section, we will introduce the construction and the improvement of the raw pairwise costs. This paper exploits the movements and appearances of the targets to construct their raw association costs with the detections in the current frame. Specifically, we establish a velocity autoregressive model \cite{dicle2013way} for the individual target and then predict the location for each object in the current frame. The Euclidean distance between each predicted position and every detection is taken as the raw motion cost for each possible pair. At the same time, we utilize the color histogram features of each target to design its raw appearance cost. Furthermore, the raw association cost for each pair is generated by coupling the raw motion cost with the raw appearance cost. In the section B, we have obtained the maximum match set for each possible pair, which can help us modify the raw association cost matrix as the Eq.(\ref{eq5}).

The appearance model would undoubtedly become ambiguous and incapable when the targets have almost the same manners. Similarly, the motion model would become unstable and powerless when the video frames are captured by a moving camera. Correspondingly, the different raw pairwise costs made of appearance and motion costs would be almost the same level. Such resemblance would inevitably make the data association assignment more ambiguous and challenging. However, with the operation of each raw pairwise cost using Eq. (\ref{eq5}), we can find that the more the number of match pairs in each match set, the lower the modified cost and vice versa. This is because the size of match set for each available pair is positively related to the structural similarity between the detections in the current frame and the historical tracklets under the condition of a fixed pair association. Obviously, the lager the size of the match set, the more likely the detection and the target in the fixed pair to be associated with each other, thus the lower their association cost and vice versa. Therefore, this modification using structural penalty makes the raw association cost, which might be generated from ambiguous appearance features and unreliable motion cues, more elaborate and discriminative. 

\begin{equation}
{{C}_{st}}(d_{i}^{t},T_{j}^{t-1})=\frac{{{n}_{\max }}}{n_{{{\Theta }_{ij}}}^{2}}\sum\limits_{\{d_{p}^{t},T_{q}^{t-1}\}\in {{\Theta }_{ij}}}{{{C}_{init}}(d_{p}^{t},T_{q}^{t-1})}
\label{eq5}
\end{equation}

\begin{equation}
{{n}_{\max }}=\max {{n}_{{{\Theta }_{ij}}}}
\label{eq6}
\end{equation}

Here, ${{C}_{init}}(d_{p}^{t},T_{q}^{t-1})$ denotes the raw association cost of the possible pair $\left\{ d_{p}^{t},T_{q}^{t-1} \right\}$. ${{C}_{st}}(d_{i}^{t},T_{j}^{t-1})$ is the association cost of the possible pair $\left\{ d_{i}^{t},T_{j}^{t-1}\right\}$ under structural constraints; and ${{n}_{{{\Theta }_{ij}}}}$ denotes the number of elements in the maximum match set of the possible pair $\left\{ d_{i}^{t},T_{j}^{t-1} \right\}$.

\subsection{Association assignment}

Data association aims to find the optimal assignment event, which corresponds to the minimum cost function, between detections and objects. In this paper, we apply the generalized linear assignment algorithm \cite{dicle2013way} to solve the data association assignment.

%\begin{itemize}
%\item The word �data� is plural, not singular.
%\end{itemize}

\subsection{Prediction of missing targets}

Generally, the false positive and false negative detections, and the situation that the targets enter into or walk out from the field of view would possibly lead to no detection in the current frame associated with the certain target in the final optimal assignment event, which is called the occurrence of missing targets. Here, we propose a target prediction method based on structural constraints as well as the motion inertia. Most of the conventional prediction methods can only promise one-frame prediction accuracy as the measurement information is not updated continuously. However, we exploit both the target motion information, which is along the time axis, and the structural information between the multi-targets, which is perpendicular to the time axis, to predict the location of missing targets. Owing to the fact that the structural information between the multiple objects is continually updated, the predicted location of a certain target would not be extended stiffly along the motion inertia, which could also alleviate the effect on motion cues caused by unpredictable moving cameras. 

We denote$\Phi $as the match set in the frame $t$ and $\overline{\Phi }$ as the set of missing targets. $T_{i}^{t}(i\in \Phi )$ and $\widetilde{T_{j}^{t}}(j\in \overline{\Phi })$ are defined by the location of the object $i$ in the match set and the predicted location of the object $j$, obtained by motion cues only, in the set of missing targets respectively. $\widehat{T_{j}^{t}}(j\in \overline{\Phi })$ represents the predicted location of the object $j$ under the constraints of both structure and motion cues. The minimization is formulated by

\begin{equation}
\begin{aligned}
  & \widehat{T_{j}^{t}}\left( j\in \bar{\Phi } \right)=\arg \underset{T_{j}^{t}\left( j\in \bar{\Phi } \right)}{\mathop{\min }}\, \\ 
 & \sum\limits_{i\in \Phi \bigcup{{\bar{\Phi }}}}{{{\left\| \Delta T_{i}^{t}-\Delta T_{i}^{t-1} \right\|}^{2}}}+\sum\limits_{j\in \bar{\Phi }}{{{\left\| \widehat{T_{j}^{t}}-\widetilde{T_{j}^{t}} \right\|}^{2}}}
\end{aligned}
\label{eq7}
\end{equation}

The extreme point of the cost function in Eq. (\ref{eq7}) is the optimal predicted location of missing targets. We set a certain time window, in which we continuously predict the location of a missing target until the target appears again. If a target does not show up in the whole time window, it would be considered as an end of a trajectory. 

\section{experiment}

In this section, we will show the performance of the proposed method on the public data set 2DMOT2015 \cite{leal2015motchallenge}, as well as the performance of other multi-object tracking approaches. It is worth mentioning that the proposed heuristic search for structure constraint could largely diminish the association noise caused by moving cameras and similar appearances, and thus to make the association cost more elaborate and discriminative. It is a universal framework, in which the initial cost can be arbitrarily designed. In this paper, we establish a velocity autoregressive model for each target to obtain its motion cost, and extract the color histogram feature of the individual target to construct its appearance cost. In addition, we use the generalized linear assignment algorithm to solve the data association. For the termination of the historical targets and the initialization of the new targets, we performed the measurements by using a fixed frame number gap (this article uses 10). We will simply terminate a target if it is not associated with any detections for 10 consecutive frames. If a detection in the current frame does not match any one of the historical trajectories, we will initialize this detection as a new target.

\begin{figure*}[thpb]
      \centering
      \graphicspath{ {images/} }     
      \includegraphics[scale=0.77]{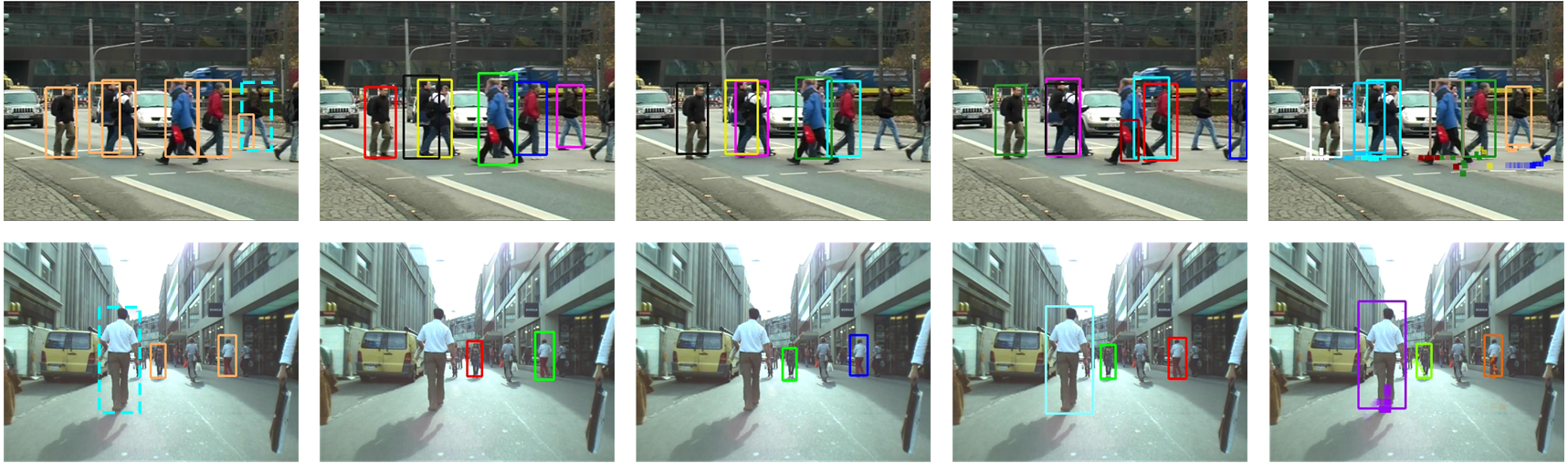}
      \caption{Recovery of the missing targets. The first column shows the pedestrian detections of a frame on the TUD-Crossing and ETH-Linthescher datasets. The bounding boxes with solid line denotes the pedestrian detections detected by ACF pedestrian detector\cite{dollar2014fast}, and the blue dash line boxes indicate the missing targets, which appeared in the past several frames. The following columns shows the tracking results of the RNN\_LSTM, the RMOT, the TC\_ODAL, and our tracker respectively. It can be seen that in the first row, the RNN\_LSTM and our tracker manage to recover the missing target. Similarly, the TC\_ODAL and our tracker manage to recover the missing object in the second row.}
      \label{figurelabel3}
   \end{figure*}

Data set 2DMOT2015 provides both training and test data sets, each of which contain 11 video sequences of different scenes and pedestrian detections obtained by Aggregate Channel Features (ACF) pedestrian detector \cite{dollar2014fast}. To evaluate of the performance of trackers, we adopt the widely used MOT evaluation metrics \cite{bernardin2008evaluating}, in which Multiple Object Tracking Accuracy (MOTA) is regarded as a comprehensive evaluation considering false detection, missed test and IDs. Multiple Object Tracking Precision (MOTP) measures the misalignment between the annotated and the predicted bounding boxes. For both MOTA and MOTP, a higher value represents a better performance. In addition, we also used other evaluation metrics like Mostly Tracked Targets (MT), Mostly Lost Targets (ML), FAF (The average number of false alarms per frame), False Positives (FP), False Negatives (FN), ID Sw. and so forth.

In the test datasets, the cameras keep moving in five scenes, including ADL-Rundle-1, ETH-Crossing, ETH-Jelmoli, ETH-Linthescher, and KITTI-19. The performance of the proposed method on these datasets is shown in table \ref{table1}. We also compared our method with other online trackers, including RMOT \cite{yoon2015bayesian}, TC\_ODAL \cite{bae2014robust}, RNN\_LSTM \cite{milan2017online}. The RMOT constructs a relative motion network to deal with the unexpected camera motion. The TC\_ODAL pay much attention to learning discriminative appearance features of targets. The RNN\_LSTM propose an end-to-end learning approach for online multi-object tracking. It can be seen that the proposed method in this paper achieves best or second best performance in MOTA as well as FP in the video frames captured by moving cameras. This is because the moving cameras would inevitably make the objects’ motion cues unstable and unpredictable. As a result, the association cost computed from motion model would become much more ambiguous and less discriminative. However the internal structure among objects in adjacent frames would basically remain constant and stable, which could be perfectly used to compensate the undermining of motion cues. The RMOT also uses the structural constraints among objects. However, the performance of RMOT is worse than our method except for the dataset ETH-Jelmoli.

\begin{table}[h]
\newcommand{\tabincell}[2]{\begin{tabular}{@{}#1@{}}#2\end{tabular}}
\caption{Comparison to other online trackers on the MOT Challenge dataset with moving cameras (pedestrian sequences)}
\scriptsize
\label{table1}
\begin{center}
\begin{tabular}{p{0.98cm}<{\centering}|p{1.1cm}<{\centering}|p{0.5cm}<{\centering}|p{0.5cm}<{\centering}|p{0.3cm}<{\centering}|p{0.3cm}<{\centering} |p{0.36cm}<{\centering} | p{0.36cm}<{\centering} | p{0.2cm}<{\centering}}

\hline
Dataset & Method & MOTA & MOTP & MT & ML & FP & FN & ID \\
\hline
\multirow{4}{*}{\tabincell{c}{ADL-\\Rundle-1}} & RMOT&	-1.3&	69.7& \textbf{21.9}&	21.9& 4790&	\textbf{4541}&	96\\
\ &TC\_ODAL& -0.2& 69.9&	15.6& \textbf{18,8}&	4476& 4782&	\textbf{70}\\
\ &RNN\_LSTM& -2.2& 69.9& 18.8& 25.0& 4213& 5058& 241\\
\ &Ours& \textbf{11.5}& \textbf{71.8}& 15.6& 28.1& \textbf{2812}& 5303&	124\\
\hline
\hline
\multirow{4}{*}{\tabincell{c}{ETH-\\Crossing}}& RMOT& 16.8&	73.6& \textbf{3.8}&	76.9& \textbf{11}& 813& 10\\
\ &TC\_ODAL& 16.3& 73.6&	0.0& 73.1& 69& 770&	\textbf{1}\\
\ &RNN\_LSTM& 21.1& \textbf{75.5}& 0.0& 57.7& 27& 757& 7\\
\ &Ours& \textbf{23.8}& \textbf{74.5}& 0.0& \textbf{53.8}& 26& \textbf{718}& 20\\
\hline
\hline
\multirow{4}{*}{\tabincell{c}{ETH-\\Jelmoli}}& RMOT& \textbf{40.4}& 71.3& \textbf{17.8}& 33.3& \textbf{263}& \textbf{1219}& 29\\
\ &TC\_ODAL& 31.2& 72.0&	13.3& 33.3& 495& 1227&	\textbf{23}\\
\ &RNN\_LSTM& 34.8& \textbf{73.3}& \textbf{17.8}& \textbf{28.9}& 314& 1280& 59\\
\ &Ours& 35.2 & \textbf{73.3}& 15.6& 33.3& 289& 1281&	 73\\
\hline
\hline
\multirow{4}{*}{\tabincell{c}{ETH-\\Linthescher}}& RMOT& 13.1&	71.9& 1.5& 81.2& \textbf{142}& 7589& \textbf{26}\\
\ &TC\_ODAL& 14.1& 73.3&	1.0& 78.7& 292& 7352& \textbf{26}\\
\ &RNN\_LSTM& 12.4& \textbf{74.7}& 1.5& 79.7& 164& 7609& 49\\
\ &Ours& \textbf{19.4}& 74.5& \textbf{2}& \textbf{72.1}& 157& \textbf{6908}& 131\\
\hline
\hline
\multirow{4}{*}{\tabincell{c}{KITTI-19}}& RMOT& 17.8&	65.5& 4.8& 32.3& 1198& 3117& 79\\
\ &TC\_ODAL& 12.9& 66.5& 4.8& 27.4& 1351& 3237&	\textbf{66}\\
\ &RNN\_LSTM& 17.7& \textbf{68.3}& \textbf{6.5}& 25.8& 1388& \textbf{2818}& 191\\
\ &Ours& \textbf{18.9}& 66.1& 4.8& \textbf{24.2}& \textbf{1152}& 3016& 166\\
\hline
\end{tabular}
\end{center}
\end{table}

Table \ref{table2} shows the overall performance of our method and other online trackers, including RMOT \cite{yoon2015bayesian}, TC\_ODAL \cite{bae2014robust} and RNN\_LSTM \cite{milan2017online} as well as offline trackers, including ALExTRAC \cite{bewley2016alextrac}, SegTrack \cite{milan2015joint} and DCO\_X \cite{milan2016multi}. The ALExTRAC utilizes appearance cues to learn an affinity model to estimate the data association cost. The SegTrack proposes a unified CRF model for joint tracking and segmentation of multiple targets. The DCO\_X models the data association problem as minimization of unified discrete-continuous energy function. From table \ref{table2}, we can conclude that our method achieves better performance on MOTA, FAF, ML and FP. The DCO\_X get the best performance on IDS. As an offline tracker, the DCO\_X deals with the data association assignment of detections-to-tracklets and tracklets-to-tracklets as well, so the length of each tracklet tends to get longer and the number of tracklets is smaller, which, to some extent, can lessen the number of IDs, but aggravate false positive detections. So does other trackers.

The performance of recovery of the missing targets directly affect the overall performance of a tracker. If the recovery method work well, it can cut down the false negatives and the IDs as well. Otherwise, it could barely decrease the false negatives and it could even increase the false positives. Therefore, when setting the parameters of our tracker, we adopt relatively prudential strategies as much as possible, which explains the low FPs, high IDs of our tracking results in table \ref{table2}. Nevertheless, the encouraging thing is that the FNs of our tracker are not as high as we thought and the ML of our tracker is better than other trackers, which indicate that our recovery method of the missing targets perform pretty well compared with other approaches.

\begin{table}[h]
\newcommand{\tabincell}[2]{\begin{tabular}{@{}#1@{}}#2\end{tabular}}
\caption{Comparison to other online trackers on the MOT Challenge dataset with moving cameras (pedestrian sequences)}
\scriptsize
\label{table2}
\begin{center}
\begin{tabular}{p{1.1cm}<{\centering}|p{0.6cm}<{\centering}|p{0.6cm}<{\centering}|p{0.3cm}<{\centering}|p{0.3cm}<{\centering}|p{0.3cm}<{\centering} |p{0.5cm}<{\centering} | p{0.5cm}<{\centering} | p{0.3cm}<{\centering}}

\hline
Method & MOTA & MOTP& FAF & MT & ML & FP & FN & ID \\
\hline
RMOT& 18.2&	69.6& 2.2& 5.3& 53.3& 12473&	36835 &	684\\
TC\_ODAL& 15.1& 70.5& 2.2& 3.2&55.8& 12970& 38538& 637\\
RNN\_LSTM& 19.0& 71.0& 2.0& 5.5& 45.6& 11578& \textbf{36706}& 1490\\
ALExTRAC& 17.0& 71.2& 1.6& 3.9& 52.4& 9233& 39933& 1859\\
SegTrack & 22.5& \textbf{71.7}& 1.4& \textbf{5.8}&63.9& 7890& 39020& 697\\
DCO\_X &19.6& 71.4& 1.8& 5.1& 54.9& 10652& 38232& \textbf{521}\\ 
Ours& \textbf{25.0}& 71.2& \textbf{1.3}& 5.0& \textbf{43.8}& \textbf{7645}& 36936& 1504\\

\hline
\end{tabular}
\end{center}
\end{table}

 Figure \ref{figurelabel3} shows the examples of recovery of the missing targets from several online MOT trackers, which is obtained from the MOT Benchmark \cite{leal2015motchallenge}. The first column shows the pedestrian detections on the TUD-Crossing and ETH-Linthescher dataset. The following columns shows the tracking result of the RNN\_LSTM, the RMOT, the TC\_ODAL, and our tracker respectively. The bounding boxes with solid lines in the first column represents the detections generated by ACF pedestrian detector \cite{dollar2014fast}, and the blue dash line boxes indicate the targets missing abruptly, which appeared in the past several frames. The tracking results illustrate that our tracker manage to recover the missing targets in both of the two frames. The RNN\_LSTM method and the TC\_ODAL tracker manage to recover the missing target once respectively and the RMOT fails to recover the missing targets. 

\section{CONCLUSION}

When the video sequences are captured by a moving camera, the motion of the targets will become unsteady, unpredictable and ambiguous. In this paper, we propose a new heuristic approach to search for the optimal internal structure constraints between multiple targets, which could be utilized to alleviate the ambiguities of the objects’ motion costs and thereby make each pairwise association cost much more elaborate and discriminative. Furthermore, we models the assignment of recovery missing targets as minimization of a cost function constructed from both motion and structure cues. Experimental results show that the proposed method achieves encouraging performance on the MOT Challenge dataset. In future work, we intend to establish a dynamic model for the structural constraint and thus make it more stable and predictable.

%\section*{APPENDIX}


\section*{ACKNOWLEDGMENT}

This work was supported by the National Natural Science Foundation of China under Grant Nos: 61273366 and 61231018 and the program of introducing talents of discipline to university under grant no: B13043.

%%%%%%%%%%%%%%%%%%%%%%%%%%%%%%%%%%%%%%%%%%%%%%%%%%%%%%%%%%%%%%%%%%%%%%%%%%%%%%%%

\bibliographystyle{IEEEtran}

\bibliography{root}

\begin{thebibliography}{10}
\providecommand{\url}[1]{#1}
\csname url@samestyle\endcsname
\providecommand{\newblock}{\relax}
\providecommand{\bibinfo}[2]{#2}
\providecommand{\BIBentrySTDinterwordspacing}{\spaceskip=0pt\relax}
\providecommand{\BIBentryALTinterwordstretchfactor}{4}
\providecommand{\BIBentryALTinterwordspacing}{\spaceskip=\fontdimen2\font plus
\BIBentryALTinterwordstretchfactor\fontdimen3\font minus
  \fontdimen4\font\relax}
\providecommand{\BIBforeignlanguage}[2]{{%
\expandafter\ifx\csname l@#1\endcsname\relax
\typeout{** WARNING: IEEEtran.bst: No hyphenation pattern has been}%
\typeout{** loaded for the language `#1'. Using the pattern for}%
\typeout{** the default language instead.}%
\else
\language=\csname l@#1\endcsname
\fi
#2}}
\providecommand{\BIBdecl}{\relax}
\BIBdecl

\bibitem{leal2015motchallenge}
L.~Leal-Taix{\'e}, A.~Milan, I.~Reid, S.~Roth, and K.~Schindler, ``Motchallenge
  2015: Towards a benchmark for multi-target tracking,'' \emph{arXiv preprint
  arXiv:1504.01942}, 2015.

\bibitem{solera2015learning}
F.~Solera, S.~Calderara, and R.~Cucchiara, ``Learning to divide and conquer for
  online multi-target tracking,'' in \emph{Proceedings of the IEEE
  International Conference on Computer Vision}, 2015, pp. 4373--4381.

\bibitem{bewley2016alextrac}
A.~Bewley, L.~Ott, F.~Ramos, and B.~Upcroft, ``Alextrac: Affinity learning by
  exploring temporal reinforcement within association chains,'' in
  \emph{Robotics and Automation (ICRA), 2016 IEEE International Conference
  on}.\hskip 1em plus 0.5em minus 0.4em\relax IEEE, 2016, pp. 2212--2218.

\bibitem{bae2014robust}
S.-H. Bae and K.-J. Yoon, ``Robust online multi-object tracking based on
  tracklet confidence and online discriminative appearance learning,'' in
  \emph{Proceedings of the IEEE conference on computer vision and pattern
  recognition}, 2014, pp. 1218--1225.

\bibitem{dicle2013way}
C.~Dicle, O.~I. Camps, and M.~Sznaier, ``The way they move: Tracking multiple
  targets with similar appearance,'' in \emph{Proceedings of the IEEE
  International Conference on Computer Vision}, 2013, pp. 2304--2311.

\bibitem{andriyenko2011multi}
A.~Andriyenko and K.~Schindler, ``Multi-target tracking by continuous energy
  minimization,'' in \emph{Computer Vision and Pattern Recognition (CVPR), 2011
  IEEE Conference on}.\hskip 1em plus 0.5em minus 0.4em\relax IEEE, 2011, pp.
  1265--1272.

\bibitem{milan2014continuous}
A.~Milan, S.~Roth, and K.~Schindler, ``Continuous energy minimization for
  multitarget tracking,'' \emph{IEEE transactions on pattern analysis and
  machine intelligence}, vol.~36, no.~1, pp. 58--72, 2014.

\bibitem{milan2016multi}
A.~Milan, K.~Schindler, and S.~Roth, ``Multi-target tracking by
  discrete-continuous energy minimization,'' \emph{IEEE transactions on pattern
  analysis and machine intelligence}, vol.~38, no.~10, pp. 2054--2068, 2016.

\bibitem{possegger2014occlusion}
H.~Possegger, T.~Mauthner, P.~M. Roth, and H.~Bischof, ``Occlusion geodesics
  for online multi-object tracking,'' in \emph{Proceedings of the IEEE
  Conference on Computer Vision and Pattern Recognition}, 2014, pp. 1306--1313.

\bibitem{collins2012multitarget}
R.~T. Collins, ``Multitarget data association with higher-order motion
  models,'' in \emph{Computer Vision and Pattern Recognition (CVPR), 2012 IEEE
  Conference on}.\hskip 1em plus 0.5em minus 0.4em\relax IEEE, 2012, pp.
  1744--1751.

\bibitem{liu2013tracking}
J.~Liu, P.~Carr, R.~T. Collins, and Y.~Liu, ``Tracking sports players with
  context-conditioned motion models,'' in \emph{Proceedings of the IEEE
  Conference on Computer Vision and Pattern Recognition}, 2013, pp. 1830--1837.

\bibitem{yang2012online}
B.~Yang and R.~Nevatia, ``An online learned crf model for multi-target
  tracking,'' in \emph{Computer Vision and Pattern Recognition (CVPR), 2012
  IEEE Conference on}.\hskip 1em plus 0.5em minus 0.4em\relax IEEE, 2012, pp.
  2034--2041.

\bibitem{milan2015joint}
A.~Milan, L.~Leal-Taix{\'e}, K.~Schindler, and I.~Reid, ``Joint tracking and
  segmentation of multiple targets,'' in \emph{Proceedings of the IEEE
  Conference on Computer Vision and Pattern Recognition}, 2015, pp. 5397--5406.

\bibitem{kim2016cdt}
H.-U. Kim and C.-S. Kim, ``Cdt: Cooperative detection and tracking for tracing
  multiple objects in video sequences,'' in \emph{European Conference on
  Computer Vision}.\hskip 1em plus 0.5em minus 0.4em\relax Springer, 2016, pp.
  851--867.

\bibitem{tang2014detection}
S.~Tang, M.~Andriluka, and B.~Schiele, ``Detection and tracking of occluded
  people,'' \emph{International Journal of Computer Vision}, vol. 110, no.~1,
  pp. 58--69, 2014.

\bibitem{yoon2015bayesian}
J.~H. Yoon, M.-H. Yang, J.~Lim, and K.-J. Yoon, ``Bayesian multi-object
  tracking using motion context from multiple objects,'' in \emph{Applications
  of Computer Vision (WACV), 2015 IEEE Winter Conference on}.\hskip 1em plus
  0.5em minus 0.4em\relax IEEE, 2015, pp. 33--40.

\bibitem{hong2016online}
J.~Hong~Yoon, C.-R. Lee, M.-H. Yang, and K.-J. Yoon, ``Online multi-object
  tracking via structural constraint event aggregation,'' in \emph{Proceedings
  of the IEEE Conference on Computer Vision and Pattern Recognition}, 2016, pp.
  1392--1400.

\bibitem{dehghan2015target}
A.~Dehghan, Y.~Tian, P.~H. Torr, and M.~Shah, ``Target identity-aware network
  flow for online multiple target tracking,'' in \emph{Proceedings of the IEEE
  Conference on Computer Vision and Pattern Recognition}, 2015, pp. 1146--1154.

\bibitem{chari2015pairwise}
V.~Chari, S.~Lacoste-Julien, I.~Laptev, and J.~Sivic, ``On pairwise costs for
  network flow multi-object tracking,'' in \emph{Proceedings of the IEEE
  Conference on Computer Vision and Pattern Recognition}, 2015, pp. 5537--5545.

\bibitem{butt2013multi}
A.~A. Butt and R.~T. Collins, ``Multi-target tracking by lagrangian relaxation
  to min-cost network flow,'' in \emph{Proceedings of the IEEE Conference on
  Computer Vision and Pattern Recognition}, 2013, pp. 1846--1853.

\bibitem{haubold2016generalized}
C.~Haubold, J.~Ale{\v{s}}, S.~Wolf, and F.~A. Hamprecht, ``A generalized
  successive shortest paths solver for tracking dividing targets,'' in
  \emph{European Conference on Computer Vision}.\hskip 1em plus 0.5em minus
  0.4em\relax Springer, 2016, pp. 566--582.

\bibitem{jiang2007linear}
H.~Jiang, S.~Fels, and J.~J. Little, ``A linear programming approach for
  multiple object tracking,'' in \emph{Computer Vision and Pattern Recognition,
  2007. CVPR'07. IEEE Conference on}.\hskip 1em plus 0.5em minus 0.4em\relax
  IEEE, 2007, pp. 1--8.

\bibitem{huang2008robust}
C.~Huang, B.~Wu, and R.~Nevatia, ``Robust object tracking by hierarchical
  association of detection responses,'' in \emph{European Conference on
  Computer Vision}.\hskip 1em plus 0.5em minus 0.4em\relax Springer, 2008, pp.
  788--801.

\bibitem{tang2015subgraph}
S.~Tang, B.~Andres, M.~Andriluka, and B.~Schiele, ``Subgraph decomposition for
  multi-target tracking,'' in \emph{Proceedings of the IEEE Conference on
  Computer Vision and Pattern Recognition}, 2015, pp. 5033--5041.

\bibitem{dehghan2015gmmcp}
A.~Dehghan, S.~Modiri~Assari, and M.~Shah, ``Gmmcp tracker: Globally optimal
  generalized maximum multi clique problem for multiple object tracking,'' in
  \emph{Proceedings of the IEEE Conference on Computer Vision and Pattern
  Recognition}, 2015, pp. 4091--4099.

\bibitem{zhang2016tracking}
S.~Zhang, Y.~Gong, J.-B. Huang, J.~Lim, J.~Wang, N.~Ahuja, and M.-H. Yang,
  ``Tracking persons-of-interest via adaptive discriminative features,'' in
  \emph{European Conference on Computer Vision}.\hskip 1em plus 0.5em minus
  0.4em\relax Springer, 2016, pp. 415--433.

\bibitem{hamid2015joint}
S.~Hamid~Rezatofighi, A.~Milan, Z.~Zhang, Q.~Shi, A.~Dick, and I.~Reid, ``Joint
  probabilistic data association revisited,'' in \emph{Proceedings of the IEEE
  International Conference on Computer Vision}, 2015, pp. 3047--3055.

\bibitem{dollar2014fast}
P.~Doll{\'a}r, R.~Appel, S.~Belongie, and P.~Perona, ``Fast feature pyramids
  for object detection,'' \emph{IEEE Transactions on Pattern Analysis and
  Machine Intelligence}, vol.~36, no.~8, pp. 1532--1545, 2014.

\bibitem{bernardin2008evaluating}
K.~Bernardin and R.~Stiefelhagen, ``Evaluating multiple object tracking
  performance: the clear mot metrics,'' \emph{EURASIP Journal on Image and
  Video Processing}, vol. 2008, no.~1, p. 246309, 2008.

\bibitem{milan2017online}
A.~Milan, S.~H. Rezatofighi, A.~R. Dick, I.~D. Reid, and K.~Schindler, ``Online
  multi-target tracking using recurrent neural networks.'' in \emph{AAAI},
  2017, pp. 4225--4232.

\end{thebibliography}

%\begin{thebibliography}{99}

%\bibitem{c1} G. O. Young, Synthetic structure of industrial plastics (Book style with paper title and editor),� 	in Plastics, 2nd ed. vol. 3, J. Peters, Ed.  New York: McGraw-Hill, 1964, pp. 15�64.

%\end{thebibliography}

\end{document}